\documentclass[letterpaper, 10 pt, conference]{ieeeconf}  

\IEEEoverridecommandlockouts                              

\usepackage{amsmath, amssymb}
\usepackage{algorithm}
\usepackage{algpseudocode}
\usepackage{graphicx}
\title{\LARGE \bf
PhysReflect-VLA: Physical Feasibility and Self-Reflective Regulation for Reliable Vision-Language-Action Policies
}

\author{Jiayu Yang, Tao Yang, Weijun Li, Xiang Chang, Fei Chao, Changjing Shang, and Qiang Shen%
\thanks{J. Yang, T. Yang, W. Li, and F. Chao are with the Department of Artificial Intelligence, School of Informatics, Xiamen University, Xiamen, China. X. Chang, C. Shang, and Q. Shen are with the Department of Computer Science, Aberystwyth University, Aberystwyth, U.K. This work was supported by the 
Xiamen Municipal Natural Science Foundation Project (No.~3502Z202573010) 
and the Key Program of the National Natural Science Foundation of 
China Joint Fund (No.~U23A20383).}%
}

\begin{document}

\maketitle
\thispagestyle{empty}
\pagestyle{empty}

\begin{abstract}

Long-horizon robotic manipulation is highly sensitive to physically infeasible transitions, contact-induced disturbances, and the lack of effective self-correction during execution. Although Vision-Language-Action (VLA) models provide strong task grounding through multimodal learning, they typically generate actions in a feed-forward manner without explicitly checking physical feasibility or diagnosing execution errors online. We present PhysReflect-VLA, a plug-and-play execution-time reliability framework that augments VLA policies with physical feasibility evaluation and structured self-reflection in a closed-loop control pipeline. A Feasibility Operator evaluates whether candidate actions induce dynamically consistent state transitions; an Action Explanation Operator verifies transition coherence; and an LLM-based Reflection Module analyzes state discrepancies to generate corrective guidance for subsequent actions. A two-stage training procedure stabilizes feasibility modeling and integrates reflection into the control loop. Experiments on multi-stage, contact-rich real-world manipulation tasks show consistent improvements in stage-wise stability and overall task success compared with representative VLA baselines with an average gain of 5.4\%. Ablation results further indicate that feasibility checking and reflection-based correction both contribute to improved execution robustness. These results highlight the importance of embedding physical consistency checks and online self-reflection for reliable long-horizon robotic manipulation.

\end{abstract}

\section{INTRODUCTION}

Enabling reliable robotic action execution in realistic deployment constraints remains a central problem in embodied intelligence \cite{ref1,ref2,ref3}. Vision-Language-Action (VLA) models have emerged as a promising solution, integrating visual perception, linguistic reasoning, and control to enable flexible task specification and generalization across a wide range of robotic manipulation tasks \cite{ref4,ref5,ref6}. These models leverage large-scale pretraining, capturing diverse semantic priors from large datasets, which enables strong task performance across different environments and modalities \cite{ref7,ref8}. Despite these strengths, adapting pretrained VLA models to specific robotic platforms under real-world constraints,  remains a major challenge.

One key challenge lies in the absence of real-time physical feasibility evaluation in most VLA systems \cite{ref9,ref10,ref11,ref12}. Traditional VLA models generate action sequences in a feed-forward manner, without explicitly assessing whether predicted actions induce physically admissible or dynamically consistent state transitions. This design can lead to catastrophic failures when actions violate contact constraints, geometric limits, or task-specific physical requirements. Even when models produce diverse long-horizon action proposals, they typically lack mechanisms to verify transition consistency before execution. As a result, small prediction errors accumulate over time, especially under distribution shifts or contact-rich interactions, leading to instability and performance degradation \cite{ref13,ref14,ref15}.

A second challenge arises from the lack of structured self-reflection during execution \cite{ref16,ref17,ref18,ref19}. While VLA policies can generate actions conditioned on visual and linguistic inputs, they generally do not evaluate discrepancies between predicted and actual outcomes after execution. Without mechanisms for diagnosing execution errors or revising subsequent decisions, the system continues following its initial plan even when deviations occur. This absence of introspective regulation prevents timely correction of mistakes and limits adaptability to unforeseen environmental changes. In long-horizon and dynamically evolving tasks, minor inconsistencies can rapidly escalate, causing irreversible failures. Particularly in contact-rich settings, execution errors compound over time, pushing the system further away from intended task states. Without self-reflective capabilities, the policy lacks the ability to detect, interpret, and correct these deviations in real time \cite{ref20}.

To address these limitations, we propose PhysReflect-VLA, a plug-and-play framework that enhances pretrained VLA policies with physical feasibility and self-reflection mechanisms. Operating at a parameter scale significantly smaller than modern VLA backbones, PhysReflect-VLA introduces real-time feasibility checks and introspective reflection mechanisms without altering pretrained weights. By evaluating the physical admissibility of proposed actions and enabling real-time correction of execution discrepancies, PhysReflect-VLA transitions VLA execution from a purely feed-forward process into a closed-loop, self-reflective control pipeline. This introspective framework enhances the robustness of action execution, enabling policies to dynamically adjust their behavior based on real-time feedback, ensuring more reliable and accurate task performance in complex, dynamic environments.

In addition to these components, PhysReflect-VLA incorporates a progressive training approach that stabilizes the integration of feasibility evaluation and self-reflection models into the pretrained VLA policy. This method ensures that the feasibility and reflection mechanisms are tightly integrated with the task-specific goals, enhancing the system's ability to adapt to dynamic and uncertain environments. These show that integrating physical feasibility and self-reflection mechanisms offers a powerful approach to enhancing the adaptability and reliability of VLA policies in real-world deployment.

The main contributions of this paper are summarized as follows:
    
    \textbf{1. PhysReflect-VLA: Physical Feasibility and Self-Reflection-Aware Policy Adaptation.}
    We introduce PhysReflect-VLA, a novel framework that augments VLA policies with mechanisms for evaluating physical feasibility and enabling introspective self-reflection during execution, allowing the policy to dynamically adjust its actions in real time based on feedback.
    
    \textbf{2. Two-Stage Training Approach for Seamless Integration.}
    We propose a two-stage training paradigm that progressively integrates physical feasibility and self-reflection models into pretrained VLA policies, ensuring stable adaptation without disrupting the original policy's capabilities.
    
    \textbf{3. Enhanced Execution Reliability in Dynamic Environments.}
    We empirically show that PhysReflect-VLA improves action execution reliability, significantly enhancing the policy's ability to perform in complex and dynamic environments, where traditional feed-forward VLA models would struggle.

\section{BACKGROUND}

Vision–Language–Action (VLA) models unify visual perception, language understanding, and control by mapping multimodal observations and task instructions to executable robot actions \cite{ref4,ref5,ref6}. Built on large vision–language backbones, recent systems such as OpenVLA, OpenVLA-OFT, and $\pi_0$ demonstrate strong semantic grounding and cross-task generalization by conditioning continuous control on high-level representations \cite{ref22,ref23,ref24}. In practice, VLA policies are often trained through imitation learning and employ autoregressive, chunked, or diffusion-based action generation strategies to improve temporal coherence and expressivity in long-horizon manipulation tasks \cite{ref13,ref14,ref25,ref26}.

Despite these advances, most VLA systems operate in a predominantly feed-forward manner at execution time. Action proposals are generated from learned correlations without explicitly evaluating whether the resulting state transitions are physically admissible or dynamically stable. Consequently, small prediction errors—especially in contact-rich or constraint-sensitive scenarios—can accumulate and lead to irreversible failures. Moreover, when discrepancies between predicted and observed states arise, existing VLA policies lack structured mechanisms for diagnosing and correcting errors online. This gap between semantic reasoning and physical grounding motivates execution-time mechanisms that incorporate explicit physical feasibility assessment and self-reflective regulation to improve long-horizon reliability \cite{ref13,ref14,ref15,ref20}.

\section{METHOD}
\label{sec:method}

\begin{figure*}[t]
    \centering
    \includegraphics[width=\textwidth]{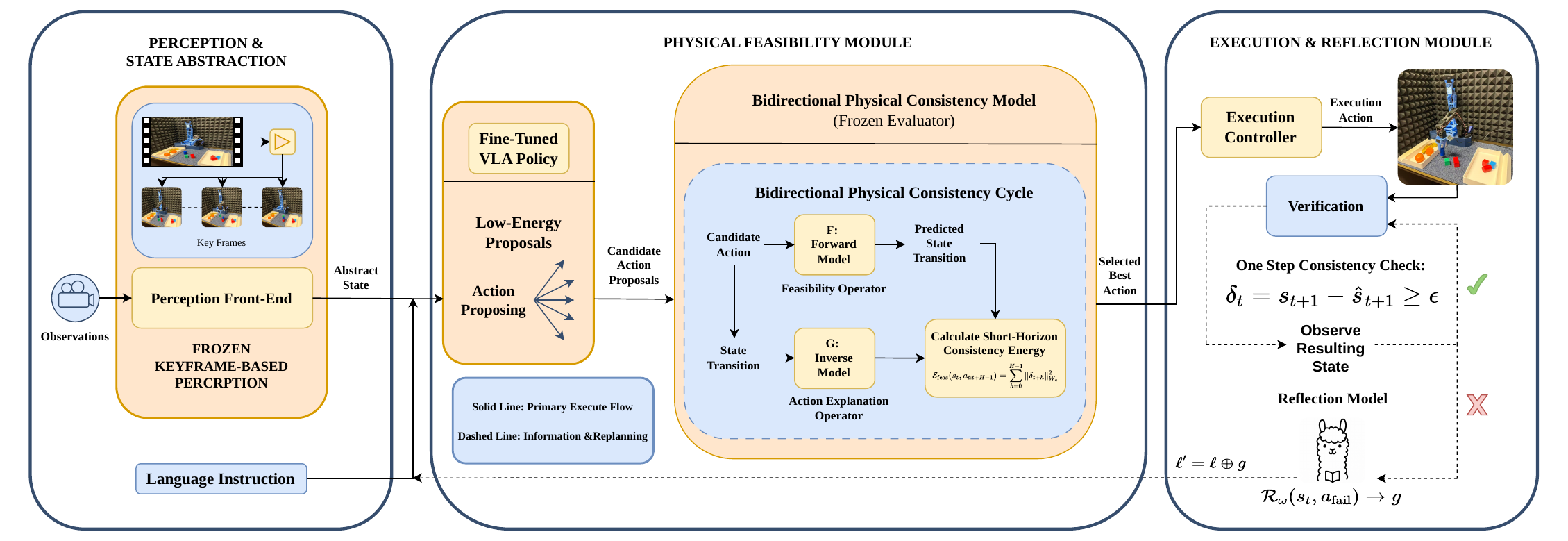}
    \caption{\textbf{Overview of PhysReflect-VLA.}
    Given visual observations and a language instruction, a base VLA policy samples multiple candidate action segments. A bidirectional feasibility evaluator, consisting of a forward transition predictor and an inverse action explainer, computes a consistency energy for each candidate and ranks them according to physical admissibility. The selected action is executed in the environment, and discrepancies between predicted and observed transitions are detected. When persistent inconsistencies arise, a reflection module generates corrective guidance that augments the original instruction and modulates subsequent action sampling. This closed-loop pipeline integrates physical feasibility evaluation and structured self-reflection to improve reliability in long-horizon manipulation.}
    \label{fig:overview}
\end{figure*}

As illustrated in Fig.~\ref{fig:overview}, We propose a reliability-oriented framework that augments a VLA policy with execution-time physical consistency modeling and structured self-reflection. Rather than treating action generation as a purely feed-forward mapping from observations and language to controls, our framework introduces a closed-loop execution structure that explicitly verifies physical admissibility and enables online behavioral correction.

At each timestep, given the current observation $o_t$ and instruction $\ell$, the VLA policy first produces one or more candidate actions (or short action segments). 
Instead of executing them directly, these candidates are evaluated in a compact abstract state space through a \textbf{bidirectional physical consistency model}. The \textbf{feasibility operator}, predicts the next state induced by a candidate action, while the \textbf{action explanation operator} reconstructs an executable action explanation from the predicted state transition. Their cycle-consistency defines a \textbf{consistency energy}, which quantifies whether the proposed action induces a physically plausible and self-consistent transition. Actions with low consistency energy are retained for execution, while inconsistent proposals are filtered or reweighted.

After execution, the observed state transition is compared with the forward prediction. If the discrepancy exceeds a tolerance threshold, a \textbf{reflection-driven correction mechanism} is activated. This module analyzes the mismatch between predicted and actual transitions and generates corrective reflection signals that condition subsequent policy inference. The policy therefore adapts its action distribution based on structured feedback about physical inconsistency, reducing error accumulation over long horizons.

By embedding physical feasibility verification and self-reflective regulation into the control cycle, the proposed framework transforms standard VLA execution into a feasibility-aware and introspection-enabled pipeline. Importantly, the design is plug-and-play: it operates alongside VLA policies without modifying their internal architecture, while substantially improving execution reliability in dynamic and contact-rich manipulation scenarios.

\subsection{Bidirectional Physical Consistency Modeling}

To evaluate whether a candidate action is physically admissible, we model consistency between actions and the state transitions they induce in a compact abstract state space.

Given the observation history $o_{1:t}$ and instruction $\ell$, a perception front-end produces a task-relevant abstract state $s_t = \mathcal{P}(o_{1:t}, \ell)$, where $s_t \in \mathcal{S}$ encodes geometric structure, object relationships, and semantic context necessary for reasoning about state transitions. The perception module $\mathcal{P}$ remains fixed, ensuring stable state abstraction during execution \cite{ref27}.

We introduce two learned mappings defined in the abstract state space:
\begin{equation}
\begin{aligned}
\mathcal{F}_\theta &: (s_t, a_t) \mapsto \hat{s}_{t+1}, \\
\mathcal{G}_\psi &: (s_t, s_{t+1}) \mapsto \hat{a}_t,
\end{aligned}
\end{equation}
where the feasibility operator, or the forward model $\mathcal{F}_\theta$ predicts the next abstract state induced by an action, and the action explanation operator, or the inverse model $\mathcal{G}_\psi$ reconstructs an executable action explanation from a state transition.

The forward model captures transition predictability under learned sensorimotor dynamics, while the inverse model captures action explainability. 
An action is considered physically consistent only if its predicted consequence admits a coherent inverse reconstruction within the same representation space.

Because manipulation tasks are dynamic and often contact-sensitive, we evaluate consistency over a short horizon $H$. 
Given an action segment $a_{t:t+H-1}$, we roll out predicted states recursively:
\begin{equation}
\hat{s}_{t+h+1} = \mathcal{F}_\theta(\hat{s}_{t+h}, a_{t+h}), 
\quad h = 0,\dots,H-1,
\end{equation}
with $\hat{s}_t = s_t$.

For each predicted transition, the inverse model reconstructs:
\begin{equation}
\hat{a}_{t+h} = \mathcal{G}_\psi(\hat{s}_{t+h}, \hat{s}_{t+h+1}),
\end{equation}
and we compute the reconstruction discrepancy
\begin{equation}
\delta_{t+h} = a_{t+h} - \hat{a}_{t+h}.
\end{equation}

We define the short-horizon consistency energy:
\begin{equation}
\mathcal{E}_{\mathrm{feas}}(s_t, a_{t:t+H-1}) 
= 
\sum_{h=0}^{H-1} \|\delta_{t+h}\|_{W_a}^2,
\end{equation}
where $W_a \succeq 0$ weights action dimensions.

Low consistency energy indicates that the proposed action segment induces transitions that are both dynamically predictable (forward-consistent) and self-explainable (inverse-consistent). 
We therefore interpret $\mathcal{E}_{\mathrm{feas}}$ as an execution-time measure of physical feasibility.

During deployment, candidate actions are ranked or filtered according to their consistency energy, enabling explicit suppression of physically inconsistent transitions before execution.

\subsection{Reflection-Guided Resampling}

While the feasibility module filters physically inconsistent actions, it does not provide explicit guidance on how a failed action should be corrected. In long-horizon manipulation tasks, simply rejecting infeasible candidates may lead to repeated sampling of similar failure modes. To address this limitation, we introduce a reflection mechanism that generates corrective guidance when execution discrepancies are detected.

During execution, the feasibility backbone predicts the next state $\hat{s}_{t+1}$ induced by a candidate action. After execution, the system observes the actual transition $s_{t+1}$ and computes a discrepancy $\delta_t = s_{t+1} - \hat{s}_{t+1}$.

When the discrepancy exceeds a predefined tolerance, the system identifies the action as a failure case and constructs a failure context consisting of $(s_t, a_{\text{best}}, \delta_t, \ell)$, where $s_t$ denotes the current state, $a_{\text{best}}$ is the candidate action with the lowest feasibility energy among the sampled proposals, and $\ell$ is the task instruction.

We introduce a lightweight reflection module (Reflector) that maps the failure context to a corrective guidance token: $\mathcal{R}_\omega :
(s_t, a_{\text{best}}, \delta_t, \ell) \rightarrow g$. The reflection token $g$ describes how the action strategy should be adjusted. Typical guidance includes suggestions such as reducing contact force, modifying approach direction, or adjusting motion magnitude. The generated reflection token is appended to the original instruction to form an augmented command $\ell' = \ell \oplus g $. Conditioned on this updated instruction, the VLA policy resamples candidate actions: $a_{t:t+H-1}^{(i)} \sim \Phi_\phi(\cdot \mid s_t, \ell')$. 

This process encourages the policy to explore action strategies that explicitly address the previously detected failure mode. In addition to instruction augmentation, the reflection signal can optionally adjust sampling parameters such as temperature, candidate set size, or action horizon length. These temporary adjustments allow the controller to increase exploration when failures are detected while maintaining efficient sampling during normal execution. Through this mechanism, reflection transforms execution discrepancies into structured corrective signals, enabling the policy to adapt its behavior online without modifying the underlying model parameters.

\subsection{Training Pipeline}

Our training procedure consists of two major phases:
(1) progressively establishing a calibrated bidirectional feasibility backbone, and (2) learning reflection-guided adaptive resampling on top of the stabilized backbone. We begin by describing data collection. All data are collected in real robots. Each rollout produces transition tuples $(s_t, a_t, s_{t+1})$, where $s_t = \mathcal{P}(o_{1:t}, \ell)$ is the abstract state extracted by the frozen perception module.

We construct:

\begin{itemize}
    \item $\mathcal{D}_{\text{mix}}$: a mixed dataset containing high-quality executions and policy-generated transitions;
    \item $\mathcal{D}_{\text{hq}}$: a high-precision dataset dominated by successful executions;
    \item $\mathcal{D}_{\text{fail}}$: discrepancy-triggered failure samples collected during feasibility-only deployment.
\end{itemize}

These datasets support progressive stabilization of feasibility modeling and reflection learning.

The feasibility backbone is trained in a progressively structured manner to ensure both compatibility with the policy distribution and high-fidelity physical calibration. We first jointly optimize the VLA policy $\Phi_\phi$ and the bidirectional feasibility models
\[
\mathcal{F}_\theta : (s_t, a_t) \rightarrow \hat{s}_{t+1}, \quad
\mathcal{G}_\psi : (s_t, s_{t+1}) \rightarrow \hat{a}_t,
\]
on the mixed dataset $\mathcal{D}_{\text{mix}}$.

The joint objective is defined as:
\begin{equation}
\mathcal{L}_{\text{joint}}
=
\mathcal{L}_{\text{policy}}
+
\lambda_1 \mathcal{L}_{\text{dyn}}
+
\lambda_2 \mathcal{L}_{\text{inv}}
+
\lambda_3 \mathcal{L}_{\text{cyc}},
\end{equation}
where $\mathcal{L}_{\text{dyn}}$, $\mathcal{L}_{\text{inv}}$, and $\mathcal{L}_{\text{cyc}}$ are defined as:
\begin{align}
\mathcal{L}_{\text{dyn}} &= 
\mathbb{E}\| s_{t+1} - \mathcal{F}_\theta(s_t,a_t) \|^2, \\
\mathcal{L}_{\text{inv}} &= 
\mathbb{E}\| a_t - \mathcal{G}_\psi(s_t,s_{t+1}) \|^2, \\
\mathcal{L}_{\text{cyc}} &= 
\mathbb{E}\| a_t - \mathcal{G}_\psi(s_t,\mathcal{F}_\theta(s_t,a_t)) \|^2.
\end{align}

This stage establishes a coarse but distribution-consistent sensorimotor geometry.

Next, we freeze the policy parameters $\phi$ and refine only $(\theta,\psi)$ on the high-quality dataset $\mathcal{D}_{\text{hq}}$. The objective strengthens physical consistency:
\begin{equation}
\mathcal{L}_{\text{refine}}
=
\mathcal{L}_{\text{dyn}}
+
\lambda_2 \mathcal{L}_{\text{inv}}
+
\lambda_3 \mathcal{L}_{\text{cyc}}.
\end{equation}

This stage sharpens the feasibility landscape and reduces over-smoothing induced by policy-generated noise.

Finally, we freeze $(\theta,\psi)$ and re-optimize the policy under feasibility regularization:
\begin{equation}
\mathcal{L}_{\text{align}}
=
\mathcal{L}_{\text{policy}}
+
\lambda_{\text{feas}}
\mathbb{E}_{a_{t:t+H-1}\sim\Phi_\phi}
\big[
\mathcal{E}_{\text{feas}}(s_t,a_{t:t+H-1})
\big].
\end{equation}

This aligns the policy distribution with the fixed physical consistency structure without distorting the feasibility backbone. After this stage, the feasibility module is fully calibrated and frozen for deployment.

Before introducing reflection learning, the system operates using feasibility filtering combined with simple replanning. Given the current state $s_t$ and instruction $\ell$, the VLA policy samples a set of candidate action segments: $a_{t:t+H-1}^{(i)} \sim \Phi_\phi(\cdot \mid s_t, \ell).$ Each candidate is evaluated by the feasibility backbone using the consistency energy $\mathcal{E}_{\text{feas}}$. If a candidate satisfies the feasibility threshold, the lowest-energy action segment is executed. If no candidate satisfies the feasibility criterion or the predicted transition is inconsistent, the controller triggers a replanning step. In this baseline setting, replanning simply resamples a new set of candidates from the same policy distribution: $a_{t:t+H-1}^{(i)} \sim \Phi_\phi(\cdot \mid s_t, \ell)$.

This process repeats until a feasible candidate is found or a replanning limit is reached. While such feasibility-aware replanning prevents obviously invalid actions, it does not provide explicit guidance on how the policy should adjust its behavior after repeated failures. In many contact-rich manipulation scenarios, the controller may repeatedly sample actions exhibiting similar failure patterns because the policy distribution remains unchanged. To address this limitation, we introduce reflection learning, which converts failure cases into structured corrective supervision and enables reflection-guided replanning.

After the feasibility backbone has been stabilized, we introduce reflection learning to enable corrective behavior when execution failures occur. We first collect failure instances during feasibility-aware policy execution. A failure case is identified when the discrepancy between predicted and observed transitions exceeds a predefined tolerance. Each failure record contains $(s_t, a_{\text{fail}}, \delta_t, \ell)$, where $s_t$ is the current state, $a_{\text{fail}}$ is the action that produced an undesirable transition, $\delta_t$ denotes the transition discrepancy, and $\ell$ is the task instruction.

To convert failure cases into structured corrective supervision, we employ a teacher vision–language model (VLM) to generate reflection guidance. Given a failure context $(s_t, a_{\text{fail}}, \delta_t, \ell)$, the teacher model produces a corrective instruction $g$ describing how the action should be modified. Typical guidance may include adjustments such as reducing contact strength, modifying the approach trajectory, or adjusting the manipulation height. Next, additional rollouts are performed to identify a feasible corrective action $a_{\text{succ}}$ that successfully resolves the failure scenario. 
This process produces labeled tuples $(s_t, a_{\text{fail}}, g, a_{\text{succ}})$, which form the reflection supervision dataset $\mathcal{D}_{\text{ref}}$.

Using the constructed dataset, we jointly train the reflector and the VLA policy. The reflector learns to predict corrective guidance from the failure context $\mathcal{R}_\omega(s_t, a_{\text{fail}}) \rightarrow g $, while the policy learns to generate the corrected action conditioned on the augmented instruction  $a_{\text{succ}} \sim \Phi_\phi(s_t, \ell \oplus g)$.

The resulting objective combines reflection supervision and corrective action learning:
\begin{equation}
\mathcal{L}_{\text{ref}}
=
\lambda_{\text{text}}
\mathcal{L}_{\text{text}}
+
\lambda_{\text{act}}
\mathcal{L}_{\text{act}},
\end{equation}
where $\mathcal{L}_{\text{text}}$ supervises reflection generation and $\mathcal{L}_{\text{act}}$ trains the policy to produce corrected actions under reflection guidance. By grounding reflection signals in teacher-labeled failure cases, the model learns to convert execution discrepancies into actionable corrective guidance.

\begin{algorithm}[H]
\caption{Training Procedure}
\label{alg:training}
\begin{algorithmic}[1]
\Require Base policy $\Phi_\phi$, perception module $\mathcal{P}$
\Require Datasets $\mathcal{D}_{mix}$, $\mathcal{D}_{hq}$
\Ensure Feasibility models $(\mathcal{F}_\theta,\mathcal{G}_\psi)$ and reflector $\mathcal{R}_\omega$

\State \textbf{/* Phase I: Feasibility Backbone */}

\State \textbf{Joint Pretraining}
\For{iterations}
    \State Sample $(o_{1:t},a_t,o_{t+1})$ from $\mathcal{D}_{mix}$
    \State $s_t \gets \mathcal{P}(o_{1:t}), \; s_{t+1} \gets \mathcal{P}(o_{1:t+1})$
    \State Compute forward, inverse and cycle losses
    \State Update $(\phi,\theta,\psi)$
\EndFor

\State \textbf{Evaluator Refinement}
\State Freeze policy $\phi$
\For{iterations}
    \State Sample $(s_t,a_t,s_{t+1})$ from $\mathcal{D}_{hq}$
    \State Update $(\theta,\psi)$ using feasibility losses
\EndFor

\State \textbf{Policy Alignment}
\State Freeze $(\theta,\psi)$
\For{iterations}
    \State Sample candidate actions from $\Phi_\phi$
    \State Compute feasibility energy $\mathcal{E}_{cons}$
    \State Update $\phi$ with feasibility regularization
\EndFor

\State \textbf{/* Phase II: Reflection Learning */}

\State Collect failure cases $(s_t,a_{fail},\delta_t)$
\For{each failure instance}
    \State $g \gets$ Teacher VLM annotation
    \State Find corrective action $a_{succ}$
    \State Add $(s_t,a_{fail},g,a_{succ})$ to dataset
\EndFor

\For{iterations}
    \State Train reflector to predict $g$
    \State Train policy to produce $a_{succ}$ conditioned on $\ell \oplus g$
\EndFor

\end{algorithmic}
\end{algorithm}
\noindent\textbf{Explanation.}
Algorithm~\ref{alg:training} summarizes the full training pipeline. We first progressively learn the feasibility backbone through joint pretraining, high-precision refinement of the forward/inverse models, and policy alignment under a frozen evaluator. We then construct teacher-annotated reflection data from failure cases and jointly train the Reflector and policy for reflection-guided correction.

\subsection{Execution}

At runtime, the controller operates in a closed loop. 
Given the current state $s_t$ and instruction $\ell$, the policy samples $K$ candidate action segments. 
The feasibility backbone evaluates each candidate using the consistency energy, and the lowest-energy feasible candidate is executed.

After execution, the observed transition is compared with the predicted transition. 
If the discrepancy exceeds a predefined tolerance, the Reflector generates a corrective guidance token $g$, which augments the instruction $\ell'=\ell\oplus g$ and triggers reflection-guided replanning. 
The policy then resamples candidate actions conditioned on $\ell'$.

\begin{algorithm}[H]
\caption{Execution with Feasibility Filtering and Reflection}
\label{alg:execution}
\begin{algorithmic}[1]
\Require Policy $\Phi_\phi$, perception $\mathcal{P}$
\Require Feasibility models $(\mathcal{F}_\theta,\mathcal{G}_\psi)$
\Require Reflector $\mathcal{R}_\omega$, instruction $\ell$
\Require candidate number $K$, thresholds $\tau_E,\tau_\delta$

\For{each timestep $t$}
    \State Observe state $s_t \gets \mathcal{P}(o_{1:t},\ell)$
    
    \State Sample candidates $\{a^{(i)}\}_{i=1}^{K} \sim \Phi_\phi(s_t,\ell)$
    \State Compute feasibility energies $E^{(i)}=\mathcal{E}_{\mathrm{cons}}(s_t,a^{(i)})$
    \State $i^\star \gets \arg\min_i E^{(i)}$
    
    \If{$E^{(i^\star)} > \tau_E$}
        \State \textbf{Resample} candidates (or relax $\tau_E$)
    \EndIf
    
    \State $a_{\text{best}} \gets a^{(i^\star)}$
    \State Execute $a_{\text{best}}$
    
    \State Observe next state $s_{t+1}$
    \State $\hat{s}_{t+1} \gets \mathcal{F}_\theta(s_t,a_{\text{best}})$
    \State $\delta_t \gets s_{t+1}-\hat{s}_{t+1}$
    
    \If{$\|\delta_t\|>\tau_\delta$}
        \State $g \gets \mathcal{R}_\omega(s_t,a_{\text{best}},\delta_t,\ell)$
        \State $\ell \gets \ell \oplus g$
    \EndIf
\EndFor
\end{algorithmic}
\end{algorithm}
\noindent\textbf{Explanation.}
Algorithm~\ref{alg:execution} describes the execution-time control loop. 
At each step, the policy samples candidate actions which are filtered and ranked by feasibility energy; infeasible sets trigger simple replanning via resampling. 
When large transition discrepancies are detected, the Reflector generates a guidance token that augments the instruction and enables reflection-guided replanning.

\section{EXPERIMENTATION}

\label{sec:experiments}

Our experiments aim to evaluate whether integrating physical feasibility evaluation and reflection-guided replanning can improve the execution reliability of VLA policies in long-horizon manipulation tasks. Specifically, we investigate three questions: (1) whether feasibility-aware action evaluation can reduce physically infeasible or unstable actions during execution, (2) whether reflection-based guidance can help recover from repeated failures caused by discrepancies between predicted and observed state transitions, and (3) whether the combination of these mechanisms improves overall task success in multi-stage manipulation scenarios.

While preliminary validation was conducted in simulation environments during the development stage, the primary focus of this work is real-world robotic manipulation. All main experimental results reported in this section are obtained from real-robot experiments designed to evaluate the reliability and robustness of the proposed execution pipeline.

\begin{table}[t]
\centering
\caption{Real-robot task success rates (\%). 
T.B.: Table-Bussy, D.C.: Drawer-Cycle, L.O.: Lid-Open, 
S.I.: Shelf-Insert, P.A.: Part-Assembly. 
S: scratch; FT: fine-tuned; OVLA: OpenVLA; Phys: PhysReflect-VLA.}
\label{tab:main_results}
\footnotesize
\setlength{\tabcolsep}{3pt}
\begin{tabular}{lcccccc}
\hline
\textbf{Method} & \textbf{T.B.} & \textbf{D.C.} & \textbf{L.O.} & \textbf{S.I.} & \textbf{P.A.} & \textbf{Avg.} \\
\hline
DP-S       & 70.0 & 81.0 & 78.0 & 59.0 & 74.0 & 72.4 \\
ACT-S      & 73.0 & 86.0 & 83.0 & 70.0 & 75.0 & 77.4 \\
RT-1-FT    & 67.0 & 73.0 & 71.0 & 61.0 & 65.0 & 67.4 \\
Octo-FT    & 71.0 & 73.0 & 72.0 & 67.0 & 69.0 & 70.4 \\
OVLA-FT    & 73.0 & 79.0 & 77.0 & 70.0 & 72.0 & 74.2 \\
Phys-OVLA  & 75.0 & 84.0 & 83.0 & 76.0 & 80.0 & 79.6 \\
OVLA-OFT   & 84.0 & 89.0 & 80.0 & 77.0 & 80.0 & 82.0 \\
Phys-OFT   & \textbf{86.0} & \textbf{91.0} & \textbf{84.0} & \textbf{79.0} & \textbf{85.0} & \textbf{85.0} \\
\hline
\end{tabular}
\end{table}


\subsection{Experimental Setup}

Our method was evaluated on five long-horizon manipulation tasks, executed primarily on a real robotic platform. Preliminary validation of the feasibility and reflection modules was first conducted in simulation during development, while the main results reported in this section are obtained from real-world experiments. 

The tasks are designed to cover both semantic reasoning and physical feasibility constraints in multi-stage manipulation: \textbf{Table-Bussy (T.B.)}: This task requires clearing a cluttered tabletop and placing a specified target object into a designated container. The robot must interpret the instruction, select relevant objects, and avoid collisions with surrounding items while completing the task. \textbf{Drawer-Cycle (D.C.)}: The robot aligns with a drawer handle, opens the drawer, places an object inside, and closes the drawer. This task involves contact-rich manipulation and requires stable control during both pulling and pushing motions. \textbf{Lid-Open (L.O.)}: The robot opens a container lid, retrieves an object from inside the container, and closes the lid afterward. Successful completion requires correct stage sequencing and reliable grasping under partial occlusion. \textbf{Shelf-Insert (S.I.)}: The robot inserts an object into a constrained slot on a shelf. This task emphasizes geometric alignment and physical feasibility during contact-based insertion. \textbf{Part-Assembly (P.A.)}: Two components must be aligned and assembled through a press-fit operation. The robot must achieve precise alignment and controlled contact to complete the assembly.

To improve statistical reliability, each method–task pair was evaluated over 20 trials under 5 different random seeds, yielding 100 runs per task. Episode success rates are computed from pooled counts across all trials, which are reported in the result tables. All real-world experiments were conducted using a 7-DoF robotic manipulator equipped with a parallel gripper. Visual observations were captured using an RGB camera positioned above the workspace. The recorded images were resized to $256 \times 256$ pixels before being processed by the VLA policy.

\subsection{Main Results}

Table.~\ref{tab:main_results} summarizes the real-robot benchmark results across five long-horizon manipulation tasks. Overall, our approach consistently improves execution reliability over the corresponding VLA baselines. Averaged across all tasks,  adding the proposed feasibility and reflection module consistently improved success rates across all five tasks. With a fine-tuning model, PhysReflect-VLA (OpenVLA) improved average success from 74.2\% to 79.6\%, and PhysReflect-VLA (OpenVLA-OFT) improved from 82.0\% to 85.0\%. In addition, reflection-guided replanning improves recovery from failure cases, reducing repeated unsuccessful attempts and stabilizing long-horizon execution.

We compared against DP \cite{ref25}, ACT \cite{ref26}, RT-1 \cite{ref13}, Octo \cite{ref14}, OpenVLA \cite{ref22}, and OpenVLA-OFT \cite{ref23}. For controlled VLA-family comparisons, PhysReflect-VLA variants used identical pretrained backbones and fine-tuning protocols as their corresponding baselines, isolating the effect of feasibility modeling. This controlled design ensures that performance differences are attributable to feasibility modeling and self-reflection mechanisms rather than backbone capacity or additional task-specific retraining.

\begin{figure*}[h]
    \centering
    \includegraphics[width=\linewidth]{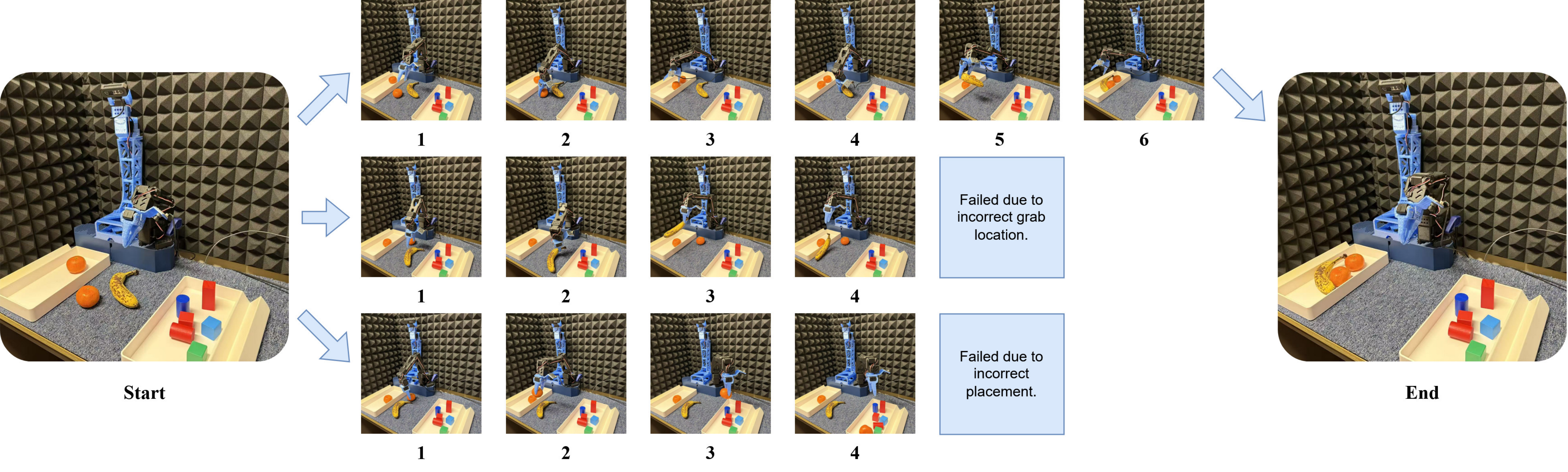}
    \caption{\textbf{Qualitative comparison on a long-horizon Table-Bussy rollout.} Starting from the same initial scene (left), the baseline VLA executes raw action proposals and fails due to physically/semantically inconsistent transitions, e.g., an incorrect grab location (middle row) or an incorrect placement (bottom row). In contrast, PhysReflect-VLA evaluates sampled candidates with the bidirectional consistency energy and preferentially executes proposals that induce physically admissible and interpretable state transitions, yielding a successful multi-step rollout (top row) and the desired final scene (right).}
    \label{fig:placeholder}
\end{figure*}

As illustrated in Fig.~\ref{fig:placeholder}, the baseline executes raw action proposals and fails when an early-stage decision leads to an inconsistent transition (e.g., an incorrect grasp location or an incorrect placement). In contrast, PhysReflect-VLA ranks sampled candidates using the bidirectional consistency energy and preferentially executes actions that induce physically admissible and interpretable transitions, resulting in a successful long-horizon rollout.

\subsection{Ablation Studies}
\label{sec:ablation}

We conduct ablation studies to disentangle the contributions of (1) bidirectional feasibility evaluation and (2) reflection-guided replanning. All ablations use the same VLA backbone and the same fine-tuning protocol as the corresponding baseline to ensure a controlled comparison.

We conducted ablations using the OpenVLA backbone with identical data and evaluation:
\begin{itemize}
    \item \textbf{Base VLA (Base):} the backbone policy executed in a standard feed-forward manner with simple replanning by resampling.
    \item \textbf{-Reflection (-R):} feasibility-aware action selection using the consistency energy, with simple replanning (no reflection).
    \item \textbf{-Feasibility (-F):} reflection-guided replanning enabled, but without feasibility-based filtering (i.e., reflection is triggered by discrepancy only).
    \item \textbf{-Consistency Training (-CT):} We keep the same execution-time candidate scoring and filtering using the consistency energy $\mathcal{E}_{\mathrm{cons}}$ (Sec.~3.2), but remove the cycle-consistency objective during training. Concretely, we set $\lambda_{\text{cyc}}=0$ and train the forward predictor $\mathcal{F}_\theta$ and inverse explainer $\mathcal{G}_\psi$ only with $\mathcal{L}_{\text{dyn}}$ and $\mathcal{L}_{\text{inv}}$. This ablation isolates the effect of \textbf{joint consistency alignment} between $\mathcal{F}_\theta$ and $\mathcal{G}_\psi$ while keeping the runtime feasibility scoring unchanged.
    \item \textbf{Full Method (Full):} feasibility-aware filtering combined with reflection-guided replanning.
\end{itemize}

As shown in Table~\ref{tab:ablation}, the full system achieves the best overall performance, improving average episode success from \textbf{74.2\%} (Base) to \textbf{79.6\%} (Full), i.e., \textbf{+5.4} points. Removing reflection (-R) reduces the average success to \textbf{77.0\%} (\textbf{-2.6} points), indicating that reflection-guided replanning provides consistent complementary gains, with the largest improvements on contact-sensitive tasks such as Shelf-Insert and Part-Assembly. Removing feasibility modeling (-F) further degrades performance to \textbf{75.6\%} (\textbf{-4.0} points), confirming that bidirectional feasibility evaluation is a key driver of reliability, particularly for tasks with strong physical constraints. 
Notably, removing cycle-consistency \emph{training} (-CT) causes the largest drop, decreasing the average success to \textbf{70.2\%} (\textbf{-9.4} points), even though the same execution-time consistency-energy scoring is still applied. This result shows that the effectiveness of feasibility-aware filtering critically depends on how well the forward predictor and inverse explainer are jointly aligned during training: without consistency alignment, the energy-based feasibility criterion becomes poorly calibrated and imposes a substantially lower performance ceiling.

Overall, the ablations indicate that feasibility modeling primarily improves reliability by suppressing physically unstable transitions before execution, while reflection improves robustness by enabling targeted recovery under persistent discrepancies. Together, they yield the strongest and most consistent performance across both semantic-heavy and physics-heavy long-horizon tasks.

\begin{table}[t]
\centering
\caption{Ablation study on five real-robot tasks.}
\label{tab:ablation}
\setlength{\tabcolsep}{5.5pt}
\renewcommand{\arraystretch}{1.15}
\begin{tabular}{lccccc}
\hline
Task & Base & -R & -F & -CT & \textbf{Full} \\
\hline
T.B. & 73.0\% & 74.0\% & 75.0\% & 70.0\% & \textbf{75.0\%} \\
D.C. & 79.0\% & 83.0\% & 80.0\% & 75.0\% & \textbf{84.0\%} \\
L.O. & 77.0\% & 81.0\% & 79.0\% & 72.0\% & \textbf{83.0\%} \\
S.I. & 70.0\% & 72.0\% & 71.0\% & 68.0\% & \textbf{76.0\%} \\
P.A. & 72.0\% & 75.0\% & 73.0\% & 66.0\% & \textbf{80.0\%} \\
\hline
\end{tabular}
\end{table}

\subsection{Analysis and Discussion}
\label{sec:discussion}

Two consistent patterns emerged. First, feasibility-aware execution delivered the largest and most consistent gains across tasks, indicating that explicit physical consistency checking provides a strong reliability bias during deployment. These benefits were most evident in later stages of multi-step manipulation, where contact-rich interactions and geometric constraints cause small transition errors to accumulate into irreversible failures. Second, reflection-guided replanning provided complementary improvements in semantic-heavy tasks, where simple resampling often repeats the same failure patterns; converting discrepancy signals into corrective language guidance helped the policy escape local failure modes and improved recovery.

The ablations further showed that reliable deployment required (1) bidirectional physical consistency to define a meaningful feasibility criterion, (2) a calibrated feasibility backbone that yields stable discrepancy signals at runtime, and (3) active closed-loop use of feasibility and reflection feedback for replanning; removing any component led to measurable and often substantial degradation.

\section{CONCLUSIONS}

This paper addressed long-horizon manipulation from the perspective of execution reliability for VLA policies. We proposed a framework that combines bidirectional physical feasibility modeling with reflection-guided replanning. The feasibility module evaluates whether candidate actions induce physically consistent state transitions, while the reflection mechanism converts execution discrepancies into corrective guidance that steers subsequent action sampling, forming a closed-loop and feasibility-aware execution pipeline.

Across five multi-stage manipulation tasks in the real world, the proposed approach shows consistent improvements in both stage-wise and episode success compared with representative VLA baselines. On average, our method improves episode success by \textbf{5.4\%}, while reducing repeated failure cases during contact-rich interactions. Ablation results further indicate that feasibility-aware filtering provides the primary performance gains, with reflection-guided correction offering additional robustness.

Despite these promising results, this work has several limitations. First, the current evaluation is conducted on a limited set of real-world manipulation tasks under controlled experimental setups, and broader task distributions are needed to further assess generality. Second, although the reported results show consistent gains, future work should include larger-scale evaluations with statistical confidence analysis and more direct comparisons to self-correction and reflective replanning methods under unified backbones, action spaces, and robot platforms. Future work will extend the framework to broader real-world task settings, explore probabilistic or multimodal feasibility models, and study more efficient reflection-guided adaptation under diverse failure modes and environmental disturbances.

\addtolength{\textheight}{-12cm}   




\bibliographystyle{IEEEtran}
\bibliography{refs}

@misc{ref1,
      title={The Developments and Challenges towards Dexterous and Embodied Robotic Manipulation: A Survey}, 
      author={Gaofeng Li and Ruize Wang and Peisen Xu and Qi Ye and Jiming Chen},
      year={2025},
      eprint={2507.11840},
      archivePrefix={arXiv},
      primaryClass={cs.RO},
      url={https://arxiv.org/abs/2507.11840}, 
}

@article{ref2,
  title={Embodied Intelligence: A Synergy of Morphology, Action, Perception and Learning},
  author={Huaping Liu and Di Guo and Angelo Cangelosi},
  journal={ACM Computing Surveys},
  year={2025},
  volume={57},
  pages={1 - 36},
  url={https://api.semanticscholar.org/CorpusID:276333529}
}

@misc{ref3,
      title={Aligning Cyber Space with Physical World: A Comprehensive Survey on Embodied AI}, 
      author={Yang Liu and Weixing Chen and Yongjie Bai and Xiaodan Liang and Guanbin Li and Wen Gao and Liang Lin},
      year={2025},
      eprint={2407.06886},
      archivePrefix={arXiv},
      primaryClass={cs.CV},
      url={https://arxiv.org/abs/2407.06886}, 
}

@misc{ref4,
      title={Pure Vision Language Action (VLA) Models: A Comprehensive Survey}, 
      author={Dapeng Zhang and Jing Sun and Chenghui Hu and Xiaoyan Wu and Zhenlong Yuan and Rui Zhou and Fei Shen and Qingguo Zhou},
      year={2025},
      eprint={2509.19012},
      archivePrefix={arXiv},
      primaryClass={cs.RO},
      url={https://arxiv.org/abs/2509.19012}, 
}

@misc{ref5,
      title={A Survey on Vision-Language-Action Models for Embodied AI}, 
      author={Yueen Ma and Zixing Song and Yuzheng Zhuang and Jianye Hao and Irwin King},
      year={2026},
      eprint={2405.14093},
      archivePrefix={arXiv},
      primaryClass={cs.RO},
      url={https://arxiv.org/abs/2405.14093}, 
}

@misc{ref6,
      title={Efficient Vision-Language-Action Models for Embodied Manipulation: A Systematic Survey}, 
      author={Weifan Guan and Qinghao Hu and Aosheng Li and Jian Cheng},
      year={2025},
      eprint={2510.17111},
      archivePrefix={arXiv},
      primaryClass={cs.RO},
      url={https://arxiv.org/abs/2510.17111}, 
}

@article{ref7,
      title={Survey of General End-to-End Autonomous Driving: A Unified Perspective},
      author={Yang, Yixiang and Han, Chuanrong and Mao, Runhao and others},
      journal={TechRxiv},
      year={2025},
      month={December},
      doi={10.36227/techrxiv.176523315.56439138/v1},
      url={https://doi.org/10.36227/techrxiv.176523315.56439138/v1}
}

@misc{ref8,
      title={Vision-Language-Action (VLA) Models: Concepts, Progress, Applications and Challenges}, 
      author={Ranjan Sapkota and Yang Cao and Konstantinos I. Roumeliotis and Manoj Karkee},
      year={2026},
      eprint={2505.04769},
      archivePrefix={arXiv},
      primaryClass={cs.CV},
      url={https://arxiv.org/abs/2505.04769}, 
}

@misc{ref9,
      title={World-VLA-Loop: Closed-Loop Learning of Video World Model and VLA Policy}, 
      author={Xiaokang Liu and Zechen Bai and Hai Ci and Kevin Yuchen Ma and Mike Zheng Shou},
      year={2026},
      eprint={2602.06508},
      archivePrefix={arXiv},
      primaryClass={cs.RO},
      url={https://arxiv.org/abs/2602.06508}, 
}

@misc{ref10,
      title={Replanning Human-Robot Collaborative Tasks with Vision-Language Models via Semantic and Physical Dual-Correction}, 
      author={Taichi Kato and Takuya Kiyokawa and Namiko Saito and Kensuke Harada},
      year={2026},
      eprint={2602.14551},
      archivePrefix={arXiv},
      primaryClass={cs.RO},
      url={https://arxiv.org/abs/2602.14551}, 
}

@misc{ref11,
      title={Rethinking Visual-Language-Action Model Scaling: Alignment, Mixture, and Regularization}, 
      author={Ye Wang and Sipeng Zheng and Hao Luo and Wanpeng Zhang and Haoqi Yuan and Chaoyi Xu and Haiweng Xu and Yicheng Feng and Mingyang Yu and Zhiyu Kang and Zongqing Lu and Qin Jin},
      year={2026},
      eprint={2602.09722},
      archivePrefix={arXiv},
      primaryClass={cs.RO},
      url={https://arxiv.org/abs/2602.09722}, 
}

@misc{ref12,
      title={FPC-VLA: A Vision-Language-Action Framework with a Supervisor for Failure Prediction and Correction}, 
      author={Yifan Yang and Zhixiang Duan and Tianshi Xie and Fuyu Cao and Pinxi Shen and Peili Song and Piaopiao Jin and Guokang Sun and Shaoqing Xu and Yangwei You and Jingtai Liu},
      year={2025},
      eprint={2509.04018},
      archivePrefix={arXiv},
      primaryClass={cs.RO},
      url={https://arxiv.org/abs/2509.04018}, 
}

@misc{ref13,
      title={RT-1: Robotics Transformer for Real-World Control at Scale}, 
      author={Anthony Brohan and Noah Brown and Justice Carbajal and Yevgen Chebotar and Joseph Dabis and Chelsea Finn and Keerthana Gopalakrishnan and Karol Hausman and Alex Herzog and Jasmine Hsu and Julian Ibarz and Brian Ichter and Alex Irpan and Tomas Jackson and Sally Jesmonth and Nikhil J Joshi and Ryan Julian and Dmitry Kalashnikov and Yuheng Kuang and Isabel Leal and Kuang-Huei Lee and Sergey Levine and Yao Lu and Utsav Malla and Deeksha Manjunath and Igor Mordatch and Ofir Nachum and Carolina Parada and Jodilyn Peralta and Emily Perez and Karl Pertsch and Jornell Quiambao and Kanishka Rao and Michael Ryoo and Grecia Salazar and Pannag Sanketi and Kevin Sayed and Jaspiar Singh and Sumedh Sontakke and Austin Stone and Clayton Tan and Huong Tran and Vincent Vanhoucke and Steve Vega and Quan Vuong and Fei Xia and Ted Xiao and Peng Xu and Sichun Xu and Tianhe Yu and Brianna Zitkovich},
      year={2023},
      eprint={2212.06817},
      archivePrefix={arXiv},
      primaryClass={cs.RO},
      url={https://arxiv.org/abs/2212.06817}, 
}

@misc{ref14,
      title={Octo: An Open-Source Generalist Robot Policy}, 
      author={Octo Model Team and Dibya Ghosh and Homer Walke and Karl Pertsch and Kevin Black and Oier Mees and Sudeep Dasari and Joey Hejna and Tobias Kreiman and Charles Xu and Jianlan Luo and You Liang Tan and Lawrence Yunliang Chen and Pannag Sanketi and Quan Vuong and Ted Xiao and Dorsa Sadigh and Chelsea Finn and Sergey Levine},
      year={2024},
      eprint={2405.12213},
      archivePrefix={arXiv},
      primaryClass={cs.RO},
      url={https://arxiv.org/abs/2405.12213}, 
}

@misc{ref15,
      title={From Mystery to Mastery: Failure Diagnosis for Improving Manipulation Policies}, 
      author={Som Sagar and Jiafei Duan and Sreevishakh Vasudevan and Yifan Zhou and Heni Ben Amor and Dieter Fox and Ransalu Senanayake},
      year={2025},
      eprint={2412.02818},
      archivePrefix={arXiv},
      primaryClass={cs.RO},
      url={https://arxiv.org/abs/2412.02818}, 
}

@misc{ref16,
      title={Vision-Language-Policy Model for Dynamic Robot Task Planning}, 
      author={Jin Wang and Kim Tien Ly and Jacques Cloete and Nikos Tsagarakis and Ioannis Havoutis},
      year={2025},
      eprint={2512.19178},
      archivePrefix={arXiv},
      primaryClass={cs.RO},
      url={https://arxiv.org/abs/2512.19178}, 
}

@misc{ref17,
      title={VLA-Reasoner: Empowering Vision-Language-Action Models with Reasoning via Online Monte Carlo Tree Search}, 
      author={Wenkai Guo and Guanxing Lu and Haoyuan Deng and Zhenyu Wu and Yansong Tang and Ziwei Wang},
      year={2026},
      eprint={2509.22643},
      archivePrefix={arXiv},
      primaryClass={cs.RO},
      url={https://arxiv.org/abs/2509.22643}, 
}

@misc{ref18,
      title={Reflective Planning: Vision-Language Models for Multi-Stage Long-Horizon Robotic Manipulation}, 
      author={Yunhai Feng and Jiaming Han and Zhuoran Yang and Xiangyu Yue and Sergey Levine and Jianlan Luo},
      year={2025},
      eprint={2502.16707},
      archivePrefix={arXiv},
      primaryClass={cs.RO},
      url={https://arxiv.org/abs/2502.16707}, 
}

@misc{ref19,
      title={CollabVLA: Self-Reflective Vision-Language-Action Model Dreaming Together with Human}, 
      author={Nan Sun and Yongchang Li and Chenxu Wang and Huiying Li and Huaping Liu},
      year={2025},
      eprint={2509.14889},
      archivePrefix={arXiv},
      primaryClass={cs.RO},
      url={https://arxiv.org/abs/2509.14889}, 
}

@misc{ref20,
      title={Inner Monologue: Embodied Reasoning through Planning with Language Models}, 
      author={Wenlong Huang and Fei Xia and Ted Xiao and Harris Chan and Jacky Liang and Pete Florence and Andy Zeng and Jonathan Tompson and Igor Mordatch and Yevgen Chebotar and Pierre Sermanet and Noah Brown and Tomas Jackson and Linda Luu and Sergey Levine and Karol Hausman and Brian Ichter},
      year={2022},
      eprint={2207.05608},
      archivePrefix={arXiv},
      primaryClass={cs.RO},
      url={https://arxiv.org/abs/2207.05608}, 
}

@misc{ref22,
      title={OpenVLA: An Open-Source Vision-Language-Action Model}, 
      author={Moo Jin Kim and Karl Pertsch and Siddharth Karamcheti and Ted Xiao and Ashwin Balakrishna and Suraj Nair and Rafael Rafailov and Ethan Foster and Grace Lam and Pannag Sanketi and Quan Vuong and Thomas Kollar and Benjamin Burchfiel and Russ Tedrake and Dorsa Sadigh and Sergey Levine and Percy Liang and Chelsea Finn},
      year={2024},
      eprint={2406.09246},
      archivePrefix={arXiv},
      primaryClass={cs.RO},
      url={https://arxiv.org/abs/2406.09246}, 
}

@misc{ref23,
      title={Fine-Tuning Vision-Language-Action Models: Optimizing Speed and Success}, 
      author={Moo Jin Kim and Chelsea Finn and Percy Liang},
      year={2025},
      eprint={2502.19645},
      archivePrefix={arXiv},
      primaryClass={cs.RO},
      url={https://arxiv.org/abs/2502.19645}, 
}

@misc{ref24,
      title={$\pi_0$: A Vision-Language-Action Flow Model for General Robot Control}, 
      author={Kevin Black and Noah Brown and Danny Driess and Adnan Esmail and Michael Equi and Chelsea Finn and Niccolo Fusai and Lachy Groom and Karol Hausman and Brian Ichter and Szymon Jakubczak and Tim Jones and Liyiming Ke and Sergey Levine and Adrian Li-Bell and Mohith Mothukuri and Suraj Nair and Karl Pertsch and Lucy Xiaoyang Shi and James Tanner and Quan Vuong and Anna Walling and Haohuan Wang and Ury Zhilinsky},
      year={2026},
      eprint={2410.24164},
      archivePrefix={arXiv},
      primaryClass={cs.LG},
      url={https://arxiv.org/abs/2410.24164}, 
}

@misc{ref25,
      title={Diffusion Policy: Visuomotor Policy Learning via Action Diffusion}, 
      author={Cheng Chi and Zhenjia Xu and Siyuan Feng and Eric Cousineau and Yilun Du and Benjamin Burchfiel and Russ Tedrake and Shuran Song},
      year={2024},
      eprint={2303.04137},
      archivePrefix={arXiv},
      primaryClass={cs.RO},
      url={https://arxiv.org/abs/2303.04137}, 
}

@misc{ref26,
      title={Learning Fine-Grained Bimanual Manipulation with Low-Cost Hardware}, 
      author={Tony Z. Zhao and Vikash Kumar and Sergey Levine and Chelsea Finn},
      year={2023},
      eprint={2304.13705},
      archivePrefix={arXiv},
      primaryClass={cs.RO},
      url={https://arxiv.org/abs/2304.13705}, 
}

@inproceedings{ref27,
  title        = {Towards a Dynamic Shapley Value-Based Evaluations for Autonomous Robotic Learning from Videos},
  author       = {Chang, Xiang and Chao, Fei and Copner, Nigel and Shang, Changjing and Shen, Qiang},
  booktitle    = {UKCI},
  pages        = {382--394},
  year         = {2025},
  organization = {Springer}
}

\end{document}